\documentclass{article}
\pdfoutput=1
\usepackage[numbers,sort&compress]{natbib}

\usepackage[final]{neurips_2020}

\usepackage[utf8]{inputenc} 
\usepackage[T1]{fontenc}    
\usepackage{hyperref}       
\usepackage{url}            
\usepackage{booktabs}       
\usepackage{amsfonts}       
\usepackage{nicefrac}       
\usepackage{microtype}      
\usepackage[most]{tcolorbox}
\usepackage{listings}
\usepackage[frozencache, cachedir=.]{minted}
\usepackage{multicol}
\usepackage{enumitem}
\usepackage{setspace}
\usepackage{caption}
\usepackage{makecell}
\usepackage{soul} 
\usepackage[export]{adjustbox} 
\usepackage[]{authblk}

\renewcommand\Affilfont{\normalsize\footnotesize}
\newcommand*{\affaddr}[1]{#1}
\newcommand*{\affmark}[1][*]{\textsuperscript{#1}}
\makeatletter
\renewcommand\AB@affilsepx{, \protect\Affilfont}
\makeatother

\author{
    \affmark[1]{\bfseries Jian Shi},
    \affmark[2,3]{\bfseries Edgar Riba},
    \affmark[4]{\bfseries Dmytro Mishkin},
    \affmark[3]{\bfseries Francesc Moreno-Noguer},
    \affmark[5]{\bfseries Anguelos Nicolaou} 
    \newline
    \affaddr{\affmark[1]The Chinese University of Hong Kong},
    \affaddr{\affmark[2]Universitat Autònoma de Barcelona},
    \newline
    \affaddr{\affmark[3]Institut de Robòtica i Informàtica Industrial},
    \affaddr{\affmark[4]Czech Technical University in Prague},
    \newline
    \affaddr{\affmark[5]Friedrich-Alexander University Erlangen-Nuremberg}
}

\title{Differentiable Data Augmentation with Kornia}

\begin{document}

\maketitle

\begin{abstract}
  In this paper we present a review of the Kornia~\citep{eriba2019kornia, eriba2020kornia} differentiable data augmentation (DDA) module for both for spatial (2D) and volumetric (3D) tensors. This module leverages differentiable computer vision solutions from Kornia, with an aim of integrating data augmentation (DA) pipelines and strategies to existing PyTorch components (e.g. autograd for differentiability, optim for optimization). In addition, we provide a benchmark comparing different DA frameworks and a short review for a number of approaches that make use of Kornia DDA.
  
\end{abstract}

\section{Introduction}
Data augmentation (DA) is a widely used technique to increase the variance of a dataset by applying random transformations to data examples during the training stage of a learning system. Generally, image augmentations can be divided in two groups: color space transformations that modify pixel intensity values (e.g. brightness, contrast adjustment) and geometric transformations that change the spatial locations of pixels  (e.g. rotation, flipping, affine transformations). Whilst training a neural network, DA is an important ingredient for regularization that alleviates overfitting problems~\cite{deeplearningbook2016}. An inherent limitation of most current augmentation frameworks 
is that they mostly rely on non-differentiable functions executed outside the computation graphs.


In order to 
optimize the augmentation parameters (e.g. degree of rotation)
by a specific objective function, differentiable data augmentation (DDA) is used. Earlier works like spatial Transformers \cite{SpatialTransformers2015} formulated spatial image transformations in a differentiable manner, allowing
 backpropagation through pixel coordinates by
 using weighted average of the pixel intensities.
 Recent works proposed to use DDA to improve GAN's training~\citep{zhao2020differentiable}, and to optimize augmentation policies~\citep{hataya2020meta, hataya2019faster}.
 
In this work, we present Kornia DDA
to help resarchers and professionals to quickly integrate
efficient differentiable augmentation pipelines
into their works. Our framework is based on Kornia~\citep{eriba2019kornia, eriba2020kornia}, which is an open-sourced computer vision library inspired by OpenCV~\citep{opencv} and designed to solve generic computer vision problems. Additionally, Kornia re-implemented classical Computer Vision algorithms from scratch in a differentiable manner and built on top of PyTorch~\citep{paszke2017automatic} to make use of the auto-differentiation engine to compute the gradients for complex operations. This paper shows: 1) the usability-centric API design, 2) a benchmark with a couple of state of the art libraries, and 3) practical usage examples.

\begin{figure}
  \begin{minipage}[b]{0.75\linewidth}
    \centering
        \begin{tabular}{llll}
        \toprule
        \multicolumn{4}{c}{\textbf{Color Space Augmentations}}\\
        \midrule
        Normalize& Denormalize & ColorJitter & Grayscale \\
        Solarize & Equalize & Sharpness  & MotionBlur \\
        MixUp &  CutMix \\
        \midrule
        \multicolumn{4}{c}{\textbf{2D Spatial Augmentations (on 4d tensor)}}\\
        \midrule
        CenterCrop  & Affine  & ResizedCrop &  Rotation \\
        Perspective & HorizontalFlip & VerticalFlip & Crop \\
        Erasing\\
        \midrule
        \multicolumn{4}{c}{\textbf{3D Volumetric Augmentations (on 5d tensor)}}\\
        \midrule
        CenterCrop3D & Crop3D & Perspective3D & Affine3D\\
        HorizontalFlip3D & VerticalFlip3D & DepthicalFlip3D \\
        \bottomrule
        \end{tabular}
    \par\vspace{0pt}
  \end{minipage}
  \begin{minipage}[b]{0.23\linewidth}
    \centering
    \includegraphics[width=.72\linewidth, right]{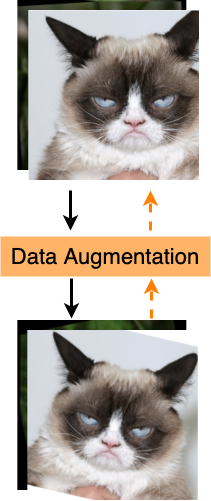}
    \par\vspace{0pt}
  \end{minipage}%
\caption{{\bf Left}: Subset of supported differentiable augmentations under Kornia 0.4.1. {\bf Right}: Our proposed scheme to represent differentiable data augmentation showing the gradients flow of the different transformations go forth and back throw the augmentations pipeline. Black arrow represents the forward pass while orange arrow represents backpropagation.}
\label{fig:test}
\end{figure}

\section{PyTorch-Oriented Design}

Kornia DDA APIs are designed following PyTorch's philosophy \cite{paszke2017automatic}, using the same logic to define neural networks as modules in order to construct complex algorithms based on differentiable building blocks.
In most existing frameworks, the DA process is a non-differentiable pre-processing step outside the computation graphs. 
While in \mintinline{python}|kornia.augmentation|, differentiable in-graph random transformations are are incorporated into the computation graph. 



\subsection{Augmentation as a Layer}

Our framework introduces augmentation layers which can be seamlessly integrated with neural network layers (e.g. Conv2D, MaxPool2D). This approach offers the following advantages:

\begin{itemize}[leftmargin=*]
    \item {\bf Automatic differentiation}. Gradients of augmentation layers could be computed whilst forward pass by taking the advantage of PyTorch autograd engine.
    \item {\bf On-device computations}. DA can be moved to any computational resource available, namely CPUs, GPUs or even TPUs. Moreover, Kornia DDA is optimized for batched data processing which can be highly accelerated by GPU and TPU. 
    \item {\bf Higher reproducibilities}. Augmentation randomness is controlled by
    PyTorch random state for the reproducible DA under the same random seed.
    In addition, DA pipeline can be serialized along with any neural networks by simply \mintinline{python}|torch.save| and \mintinline{python}|torch.load|.
    
\end{itemize}

\begin{tcolorbox}[every float=\centering, drop shadow, title=Example 1: DA Pipeline]
    \label{fig:examples:pipeline}
\begin{minted}{python}
import kornia.augmentation as K
class MyAugmentationPipeline(nn.Module):
    def __init__(self):
        super(MyAugmentationPipeline, self).__init__()
        self.mixup = K.RandomMixUp(p=1.)
        self.aff = K.RandomAffine(360, p=0.5)
        self.jitter = K.ColorJitter(0.2, 0.3, 0.2, 0.3, p=0.5)
        self.crp = K.RandomCrop((200, 200))
    def forward(self, input, label):
        input, label = self.mixup(input, label)
        input = self.crp(self.jitter(self.aff(input)))
        return input, label
aug = MyAugmentationPipeline()
\end{minted}
    \tcbsubtitle{On-device Computations}
\begin{minted}{python}
augmented = aug(images.to('cuda:0'))  # will happen in device cuda:0
augmented = aug(images.to('cuda:1'))  # will happen in device cuda:1
\end{minted}
    \tcbsubtitle{Save and Load}
\begin{minted}{python}
torch.save(aug, "./saved_da.pt")
aug_restored = torch.load("./saved_da.pt")
\end{minted}
\end{tcolorbox}

\subsection{PyTorch-Backended Optimization} \label{optimization}
Our framework provides an easy and intuitive solution to backpropagate the gradients through augmentation layers using the native PyTorch workflow. In any augmentations, \mintinline{python}|kornia.augmentation| takes \mintinline{python}|nn.Parameter| as differentiable parameters while \mintinline{python}|torch.tensor| as static parameters. The following example shows how to optimize the differentiable parameters (including brightness, contrast, saturation) of \mintinline{python}|kornia.augmentation.ColorJitter| and backpropagate the gradients based on the computed error from a loss function.

\begin{tcolorbox}[every float=\centering, drop shadow, title=Example 2: Optimizable DA]
    \label{fig:examples:optimize}
\begin{minted}{python}
import kornia.augmentation as K; import torch; import torch.nn as nn;
t = lambda x: torch.tensor(x); p = lambda x: nn.Parameter(t(x))
torch.manual_seed(42);

images = torch.tensor(img, requires_grad=True)
jitter = K.ColorJitter(
    p([0.8, 0.8]), p([0.7, 0.7]), p([0.6, 0.6]), t([0.1, 0.1]))
out = jitter(images)

loss = nn.MSELoss()(out, images)
optimizer_img = torch.optim.SGD([images], lr=1e+5) # Large lr for demo
optimizer_param = torch.optim.SGD(jitter.parameters(), lr=0.1)

loss.backward()
optimizer_img.step()
optimizer_param.step()
\end{minted}
    \tcbsubtitle{Updated Image}
    \includegraphics[width=1.\linewidth]{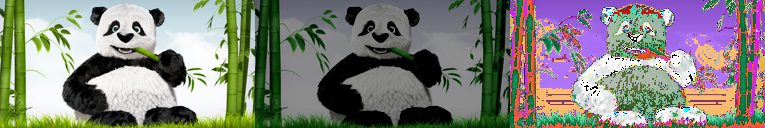}
    From left to right: the original input, augmented image and gradient-updated image.
    \tcbsubtitle{Updated Parameters}
\begin{minted}{python}
brightness -> [0.8048, 0.8363]      contrast -> [0.7030, 0.7323]
saturation -> [0.5999, 0.5976]      hue      -> [0.1000, 0.1000]
\end{minted}
\end{tcolorbox}

\section{Benchmarks}

Existing libraries such as TorchVision (based on PIL) and Albumentations~\cite{info11020125} (based on OpenCV) are optimized for CPU processing taking the advantage of multi-threading. However, our framework is optimized for GPU batch processing that runs in a synchronous manner as a precedent module for neural networks. The different design choices among those libraries determined the difference in the performance under different circumstances (e.g. hardware, batch sizes, image sizes). As we state in \autoref{tab:benchmark}, TorchVision/Albumentations show a better performance when lower computational resources are required (e.g. small image size, less images), while Kornia DDA gives a better performance when there is a CPU overhead. For the performance experiments, we used Intel Xeon E5-2698 v4 2.2 GHz (20-Core) and 4 Nvidia Tesla V100 GPUs. The code for the experiments will be publicly provided to compare against other hardware.


\begin{table}[htb]
\centering
\small
\begin{tabular}{cccc}
\toprule
 \makecell{\\Num. GPUs for}  & \multicolumn{3}{c}{\makecell{Comparison Among Different Image Sizes\\(Kornia / Albumentations / TorchVision)}}\\
Data Parallelism &  32x32 &  224x224 &  512x512 \\
\midrule
1 &  14.28 / {\bf 12.07} / 12.33 &  14.23 / {\bf 12.10} / 12.48 &  14.22 / {\bf 12.08} / 12.77 \\
\midrule
2 &  15.99 / 16.47 / {\bf 14.06} &  {\bf 12.85} / 12.93 / 13.91 &  {\bf 12.93} / 13.34 / 14.03\\
\midrule
3 &   16.61 / 17.88 / {\bf 15.21} &  {\bf 12.97} / 14.46 / 15.00 &   {\bf 13.08} / 13.96 / 15.36 \\
\midrule
4 &   16.87 / 18.99 / {\bf 15.66}  &  {\bf 13.32} / 15.38 / 15.94  &    {\bf 13.44} / 15.84 / 16.12\\
\bottomrule
\end{tabular}



\caption{\label{tab:benchmark} {\bf Speed benchmark among DA libraries.} The results are computed as the time cost (seconds) of training 1 epoch of ResNet18 using 2560 random generated faked data. Specifically, DA methods compared are RandomAffine, ColorJitter and Normalize. Batch size is 512 in all the experiments. The add-on GPU memory cost from \mintinline{python}|kornia.augmentation| is negligible. }
\end{table}

\section{Use-case examples}
\label{sec:examples}

In this section, we describe two state of the art computer vision approaches that use Kornia DDA APIs as the main backend: local feature orientation estimator and data augmentation optimization.


\subsection{Learning local feature orientation estimator with Kornia}
\begin{figure*}[htb]
\centering
\includegraphics[width=0.95\linewidth]{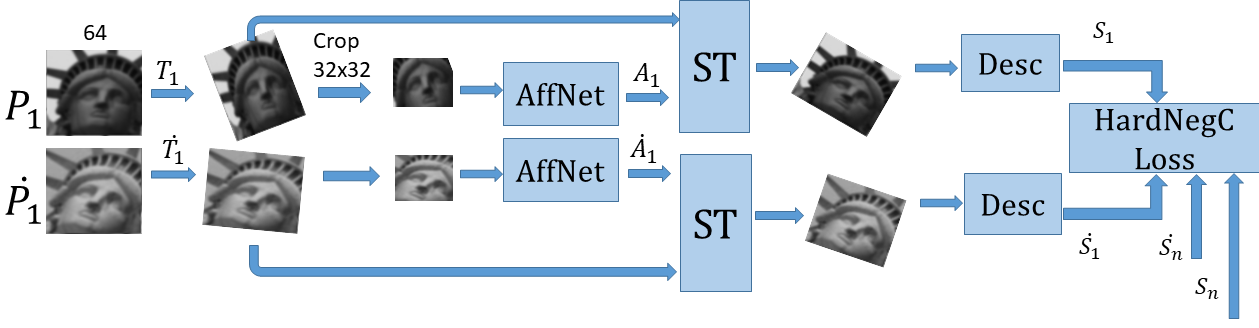}
\caption{AffNet local feature shape and orientation estimation training training, image courtesy~\cite{AffNet2018}. We re-implemented differentiable image transformations, $T_i, \dot{T}_i$, and $A_i, \dot{A}_i$ using \mintinline{python}|kornia.augmentation| functions, reducing relevant code line counts by order of magnitude.}
\label{fig:affnet-training-scheme}
\end{figure*}

An approach to learn a local feature detector is by using differentiable random spatial transformations~\cite{AffNet2018}, that is, spatial augmentation. The learned model has to predict the local patch geometry, which is then described by a local descriptor and the matching-related loss is minimized, as shown in Figure~\ref{fig:affnet-training-scheme}. Authors' implementation of all related function based on PyTorch~\cite{paszke2017automatic} \textit{grid\_sample} function takes around 600 lines of code. The same functionality can be implemented with Kornia and \mintinline{python}|kornia.augmentation| using 30 lines of code including all necessary imports. 

\subsection{Optimizable Data Augmentation with Kornia}

Designing a proper combination of DA operations is a complicated task, which often requires the specific domain knowledge. On top of the Kornia DDA module, Faster AutoAugment \cite{hataya2019faster} is implemented with an aim of learning the best augmentation policies by gradient optimization methods. Moreover, MADAO \cite{hataya2019faster} (Meta Approach to Data Augmentation Optimization) is also implemented on top of Kornia which could optimize both deep learning models and DA policies simultaneously, which effectively improved the classification performance using gradient descent.





\section{Discussion}

We presented Kornia DDA that aligned with PyTorch API design principles with a focus on usability, to perform efficient differentiable augmentation pipelines for both production and research. In addition, we inherit the differentiability property that will help researchers to explore new DDA strategies with which we believe that can change the paradigm for designing handcrafted augmentation policies.
Our future directions, will be to increase the efficiency of the different operators through the PyTorch JIT compiler and creating a generic API to perform meta-augmentation-learning.




\bibliographystyle{unsrt}
\bibliography{references}

\end{document}